\documentclass[a4paper]{spie}  

 \usepackage{lipsum} 
\usepackage{latexsym}
\usepackage{booktabs}
\usepackage{enumitem}
\usepackage{color}
\usepackage[english]{babel}
\usepackage{amsmath,amsfonts,amssymb}
\usepackage{graphicx}
\usepackage[colorlinks=true, allcolors=blue]{hyperref}
\usepackage{times}
\title{A Light Perspective for 3D Object Detection}

\author[a]{Marcelo Eduardo Pederiva}
\author[a]{José Mario De Martino}
\author[b]{Alessandro Zimmer}
\affil[a]{School of Electrical and Computer Engineering, State University of Campinas, Brazil}
\affil[b]{ AImotion Bayern, Technische Hochschule Ingolstadt, Germany}

\begin{document} 
\maketitle

\begin{abstract}
Comprehending the environment and accurately detecting objects in 3D space are essential for advancing autonomous vehicle technologies. Integrating Camera and LIDAR data has emerged as an effective approach for achieving high accuracy in 3D Object Detection models. However, existing methodologies often rely on heavy, traditional backbones that are computationally demanding. This paper introduces a novel approach that incorporates cutting-edge Deep Learning techniques into the feature extraction process, aiming to create more efficient models without compromising performance. Our model, NextBEV, surpasses established feature extractors like ResNet50 and MobileNetV2. On the KITTI 3D Monocular detection benchmark, NextBEV achieves an accuracy improvement of 2.39\%, having less than 10\% of the MobileNetV3 parameters. Moreover, we propose changes in LIDAR backbones that decreased the original inference time to 10 ms. Additionally, by fusing these lightweight proposals, we have enhanced the accuracy of the VoxelNet-based model by 2.93\% and improved the F1-score of the PointPillar-based model by approximately 20\%. Therefore, this work contributes to establishing lightweight and powerful models for individual or fusion techniques, making them more suitable for onboard implementations.
\end{abstract}

\keywords{Object Detection, Monocular detection, Sensor Fusion, Autonomous vehicles}

\section{INTRODUCTION}
\label{sec:intro}  

Ground transportation has evolved significantly, from early rudimentary cars to today's sophisticated vehicles. Integrating Machine Learning and sensory perception has enhanced vehicles' environmental understanding and promoted the development of many tasks, such as Semantic Segmentation, Line Detection, and Object Detection. Among those, 3D Object Detection stands out in mapping objects on the street by predicting their position, orientation, and size in three dimensions.

Two types of sensors are primarily used in this field: LIDAR and camera. LIDAR is widely used due to its accuracy in 3D comprehension of the environment, but it struggles to differentiate similarly shaped objects and is costly. On the other hand, camera sensors provide rich color information and excel in multi-classification tasks, but lack depth measurements and perform poorly in low light. To address these limitations, the 3D object detection field has evolved towards the fusion of both sensors, combining LIDAR's precise distance measurement with the camera's detailed visual information. LIDAR-camera fusion for 3D Object detection leads to more accurate and robust predictions. However, fusion models require larger and more complex networks, which result in high computational demand, challenging real-time responses, and onboard implementation.

The prevalent method for integrating sensor data involves passing each sensor through an individual processing path, extracting their features, converting them into a Bird-Eye-View (BEV), and then fusing both information to predict the 3D objects. In the LIDAR path, current approaches \cite{Bai2022, Li2023, Liu2022bevfusion} take advantage of the 3D LIDAR measurements and use a single architecture to extract the features and convert it into BEV directly \cite{Zhou2018, Lang2019}. On the other hand, the lack of depth measures makes the camera path challenging to predict in BEV reference. The common approach follows a sequence of two architectures with one feature extraction (Backbone) followed by a BEV conversion, which can take more than 500 milliseconds to be processed \cite{Liu2022bevfusion}. In 2023, Liu and Tang presented a solution to speed up the BEV conversion through parallelized GPU kernels, decreasing significantly the processing time of this step \cite{Liu2022bevfusion}. However, like other works \cite{Li2023, Bai2022}, their camera backbone is still based on well-known architectures that use many parameters and require high computational resources  \cite{Howard2017, Sandler2018, He2016}. As mentioned by the BEVFusion authors \cite{Liu2022bevfusion}, besides the gain in accuracy, the camera's backbone has a significant impact on increasing the model's latency.

With the focus in lightweight the processing networks, our paper revisit the camera and LIDAR path to propose updates in each one. In camera processing path, we present a new model that combines feature extraction and BEV conversion in a simple and efficient approach. This redesign considers the camera backbone's impact on model size, computational demand, and model inference time. In addition, we revisit standard LIDAR backbones to propose slight yet effective improvements to lightweight the LIDAR processing path. Lastly, we present a fusion of these optimized paths, ending up with a lightweight sensor fusion model.

In the camera path, inspired by the 3D Monocular Object Detection model MonoNext \cite{Pederiva2023}, our model leverages the latest techniques to improve the balance between lightweight, speed, and accuracy, surpassing the performance of traditional feature extraction models, like ResNet50, MobileNet, and Monocular Object detection approaches. Our approach achieves a higher accuracy using less than half of the MobileNetV2 parameters and FLOPS. Consequently, NextBEV provides a trade-off between accuracy and computational demand relevant to onboard implementations. On the other hand, in the LIDAR path, with Depthwise-based blocks, we lightweight two traditional backbones. The proposed modifications imply a reduction of 10 ms during the inference phase, which is a significant factor in engineering lighter approaches for onboard implementations seeking real-time predictions. In the end, by integrating the contributions of both processing paths into a Sensor Fusion model, we have achieved a 20\% improvement in the F1-score relative to approaches that rely on a single sensor. This integration results in a model that is not only lightweight and accurate but also effective in minimizing the occurrence of both False Positives and False Negatives in its predictions.

Our contributions can be summarized as follows:

\begin{itemize}
    \item We introduce a novel end-to-end approach that directly converts the camera image into a BEV feature tensor. The network is based on monocular detection and state-of-the-art image feature models.

    \item We analyse the LIDAR backbone models and propose modifications to decrease architecture size, computational demand, and inference time without compromising accuracy.
    
    \item We detail the implementation of our method within a combined LIDAR-Camera framework, highlighting its advantages over single-sensor solutions in terms of reliability, performance, and inference speed.
\end{itemize}


\section{Related Work}

\subsection{Camera-based 3D Object Detection}

Camera-based methods seek to estimate objects in three dimension without depth information, based only on 2D images. Stereo models rely on different perspectives of the same object, resulting in a better depth estimation of its detection in three dimensions. Many works propose various approaches to fuse camera information \cite{Chen2020}. Among them, the Stereo R-CNN \cite{Li2019} extends Faster R-CNN for stereo inputs, associating objects in left and right images to achieve high accuracies. The Pseudo-LIDAR \cite{Wang2019} converts image-based depth maps to Pseudo-LIDAR representations, enabling LiDAR-equivalent 3D detection using stereo vision. Besides superior performance due to using different perspectives, stereo models require a high computational demand to process all images, presenting a slow inference time.

On the other hand, Monocular models demonstrate the potential of deriving detailed 3D information from a single 2D image in a fast, straight approach. Although this approach does not achieve the same accuracy as stereo models, it presents solutions for 3D estimation using a lightweight, faster, and lower-cost technique. In the monocular model's field, several works present different approaches for 3D estimation \cite{Li2020,Brazil2019}. Recent contributions, like MoVi3D, advance the field with a single-stage deep architecture that utilizes geometrical information to create virtual perspectives, normalizing the appearance of objects relative to their distance \cite{Simonelli2020}. Addressing the challenges of occluded objects in Monocular 3D Object Detection, MonoPair introduces a novel approach that considers the spatial relationships between object pairs. The approach enhances detection accuracy, particularly for hard samples, and ensures runtime efficiency \cite{Chen2020monopair}. Recently, MonoNext emerged as a straightforward implementation of ConvNext-based network \cite{Liu2022} to extract features and map objects via a small spatial grid \cite{Pederiva2023}.

The majority of the methods mentioned rely on a two-step detection strategy, starting with a well-known camera backbone followed by an Image-to-BEV conversion. Such models utilize or take inspiration from established backbones in the literature, such as ConvNext \cite{Liu2022}, MobileNet \cite{Howard2017,Sandler2018}, or ResNet \cite{He2016}, before transforming features into a grid representation through another neural network. The unification of these two steps into a straight approach is proposed by the present work. Additionally, we analyze the feature extraction capability of our approach against well-known backbones and monocular models.

\subsection{LIDAR-based 3D Object Detection}

3D Object detection methods using LIDAR have evolved through various strategies, each with unique advantages and challenges. Point-based Approaches directly process the raw point cloud data from LIDAR. Pioneering this field, PointNet \cite{Charles2017} emerged as a groundbreaking framework, introducing a novel neural network approach for classification and segmentation tasks straight from point cloud data. Further methods have built upon PointNet's foundation, seeking enhancements in processing efficiency and accuracy \cite{Shi2023}. Despite the advantage of utilizing unprocessed point clouds for direct learning, these methods are often constrained by their high demand for computational resources. In contrast, Voxel-based Approaches converts the raw point cloud into a 3D voxel grid, simplifying the data representation. VoxelNet \cite{Zhou2018} led this approach with 3D Convolutional Layers to extract the features of the voxelized data, improving processing efficiency. On the other hand, the voxelization step can result in the loss of critical data details, potentially affecting the model's ability to capture relevant nuances of the data. To address the issue, other researchers \cite{Yan2018} focused on refining voxel-based methods to retain more detail and enhance overall performance.

Finally, \textbf{Hybrid-based Approaches} combine the strengths of both previous techniques. The PointPillar introduced a step to encode the point cloud into a pseudo-image \cite{Lang2019}. This hybrid approach allows for the detailed representation of point data within a structured 2D grid, which can then be efficiently processed using 2D convolutional networks. It results in a balance between computational efficiency and accuracy. Following this paradigm, other studies have adopted similar methodologies to enhance their object detection capabilities \cite{Qi2018}.

\subsection{LIDAR-Camera Fusion for 3D Object Detection}

The fusion of LIDAR and Camera data emphasizes the complementary strengths of both sensors to enhance 3D object detection. It combines LIDAR's precise depth information with the detailed color information from camera images, improving detection accuracy and robustness. The combination of both sensors typically is done in Early \cite{Chen2017} or Late stages \cite{Qi2018,Sindagi2019}. In Early Fusion, the raw sensor measurements are combined and passed through a single architecture. This process achieves fast and robust detections, sacrificing precision. On the other hand, Late Fusion processes each sensor individually and then fuses its prediction. It leads to the best accuracy, though increasing computational costs due to the parallel network architecture.

Recent proposals have introduced a Middle Fusion strategy, where data from both sensors passes through a small independent feature extractor (backbone). This fusion process is refined by a final network that fine-tunes the combined information for accurate predictions. TransFusion leverages transformer architectures to a robust fusion method that incorporates a soft-association mechanism to mitigate the effects of poor-image conditions like bad illumination and sensor misalignment \cite{Bai2022}. BEVFusion, on the other hand, moves away from traditional point-level fusion by unifying multi-modal features in a shared BEV representation space, efficiently preserving both geometric and semantic information \cite{Liu2022bevfusion}. Similar to Monocular models, the camera branch of these fusion models follows the use of a known heavy camera backbone (Swin-T or ResNet50) followed by an Image-to-BEV approach. Despite of the improve in accuracy and robustness, the camera path implement a computational demand that compromise fast detections in onboard implementations. 

\begin{figure}[t]
\centering
\includegraphics[width=0.95\textwidth]{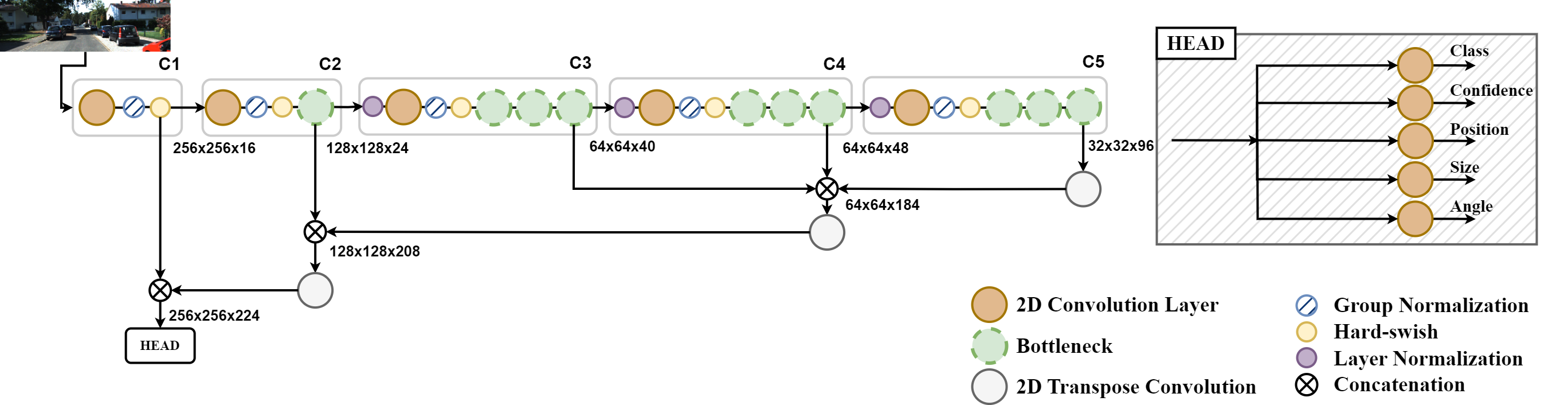}
\caption{NextBEV Architecture.}
\label{fig:NextBEV}
\end{figure}

\section{Method} 

\subsection{Camera processing path}
\label{img2bev}
We introduce NextBEV, a novel Image to Bird-Eye-View conversion model that seeks to enhance the camera processing path for Monocular detection and Sensor Fusion applications. Our method is inspired by the MonoNext model, an approach that uses complex convolution blocks to extract the image features and predict the 3D object characteristics. However, MonoNext’s high computational demand makes it difficult for onboard implementations. In contrast, MobileNet-based architectures balance size, speed, and accuracy in image feature extraction using Depth-wise Separable convolutions, fewer channels, and simpler activation functions. In light of this, our model mixes the modified MonoNext model with the latest techniques implemented by MobileNet \cite{Howard2017,Sandler2018,Howard2019}. The overall NextBEV architecture, as shown in Figure \ref{fig:NextBEV}, is structured in five cells (C1 to C5), each comprising a sequence of 3x3 kernel Convolution layer, a Group Normalization \cite{Wu2018}, and a H-swish activation function. The Convolutions use 2, 2, 2, 1, 2 strides, and 16, 24, 40, 48, 96 channels respectively. Additionally, excluding the initial segment, the subsequent four cells present a similar Depthwise Block that repeats 1 time in the second cell and 3 times in the following ones. The Depthwise Block consists of a sequence that includes a Depthwise Convolution (kernel: 3,stride: 1), Group Normalization, a Pointwise convolution to expand the channels 4 times, H-swish activation, and an additional Pointwise convolution to return to the original channels. The sequence's output is summed with the Depthwise Block input. The tensor is then upscaled with a Transpose Convolution (kernel: 3, stride: 2) and concatenated with outputs from the third and fourth segments, culminating in a $64\times64\times184$ tensor. The output undergoes further a similar upscaling and integration with the output from the second segment. Finally, it is upscaled again and concatenated with the first cell.

For the 3D predictions, the NextBEV's output is fed into a Multi-task learning head (HEAD), segmenting the input tensor into five convolutional paths with 64 channels and 1x1 kernels. Each path implements a different activation function for their prediction, with sigmoid for confidence, softmax for classification, and ReLU for position, size, and rotation. According to Qiao et al.\cite{Qiao2019}, the combination of Group Normalization (GN) \cite{Wu2018} and Weight Standardization (WS) markedly enhances performance over traditional Batch Normalization techniques. Consequently, WS is implemented across all CNN to optimize their performance.

\subsection{LIDAR processing path}
\label{lidarup}

We use PointPillar and VoxelNet methods to extract LIDAR sensor features and predict 3D object characteristics. Both methods have Convolutional Neural Network (CNN) blocks that can be optimized by current deep learning techniques. The \textbf{PointPillar} is structured with a Pillar Feature Net to convert the point cloud into a Pseudo Image. Next, this output goes to a 2D CNN built with 3x3 Conv-layers, Batch Normalization, and ReLU activation. The result is then passed to a Head to do the 3D predictions. On the other hand, \textbf{VoxelNet} follows a similar approach, beginning with the point cloud entering its feature learning Network, progressing through 3D CNN and 2D CNN built with similar blocks as PointPillar. Both 2D feature extraction methods are built in 2D CNN blocks that can be optimized with our proposed lightweight Block. Based on that, we replaced these Blocks with our Depthwise Block with the expansion of 1, Batch Normalization with Group Normalization and ReLU with H-swish activation.

\subsection{Fusion step}
\label{fusion}

With the LIDAR and Camera processing paths converted into a BEV features grid, we fuse them with the Cross-Attention (CA) approach. Inspired by Transformers, Cross-attention is a familiar method to fuse information of different types, like text and images \cite{Lee2018} and proved as a good fusion of these sensors in other works \cite{Kim2022,Li2022}. 

The features of LIDAR are used as Query, while the Camera features as Key and Value. The Attention mechanism is calculated by the multiplication between the Query tensor and the transpose of the Key tensor. This product is then normalized by the square root of the component's tensor size and passed through a softmax function, resulting in attention scores representing the correlation between features from both sensors. The attention scores are then used to weight the Value tensor, emphasizing the features relevant to both backbones. The output of the Cross-attention step is added to the LIDAR feature tensor, creating a skip connection to use the CA to emphasize its relevant features. In our approach, we use six CA heads and concatenate them into a final tensor. This output goes into a multi-task learning head (HEAD) to predict the 3D Objects' characteristics.

\section{Experiments}
\subsection{Dataset and Metrics}
\textbf{Dataset:} We benchmark our approach using the KITTI dataset \cite{Geiger2012}, which includes data from both cameras and a Velodyne HDL-64E LIDAR sensor. The dataset contains 7481 frames with Camera and LIDAR measurements for training and testing. We divided the training data into two sets: training (3741 frames) and validation (3740 frames) split \cite{Wu2022}. The dataset contains labels for different objects, like cars, vans, pedestrians, and cyclists. In this work, we evaluate the models for just detecting cars. 

\noindent\textbf{Metrics and Implementation Details:} We evaluate the model using 3D Average Precision (AP) with 40 recall thresholds. Specifically, Precision, Recall, and F1-score are calculated as follows: Precision = $\frac{TP}{(TP + FP)}$, Recall = $\frac{TP}{(TP + FN)}$, and F1-Score = $\frac{2 \times \text{Precision} \times \text{Recall}}{\text{Precision} + \text{Recall}}$. Inspired by KITTI metrics, a detection is considered correct if IoU $>$ 0.7. Additionally, we employ a 0.1 IoU threshold to verify the model's ability to recognize objects in the correct regions. Detection difficulty is categorized into three levels: Easy, Moderate, and Hard, corresponding to fully visible, partially occluded, and mostly occluded objects, respectively.

To ensure a fair comparison, inference time is averaged over 20 measurments, with a consistent computational environment and a warm-up phase. The detection range (horizontal, vertical, and depth axis) is standardized to [-40, 40; -2, 6; 0, 80] for PointPillar and Monocular models, while VoxelNet and its fusion use ranges of [-40, 40; -1, 3; 0, 70.4]. All models utilize anchors of 1.6m width, 1.5m height, and 3.9m length. Experiments were conducted on an RTX2080ti GPU, with batch sizes of 8 for Monocular models and 2 for LIDAR and Sensor fusion models. The models are trained for 180 epochs, selecting the weight based on minimum validation loss. We employ the AdamW optimizer with an initial learning rate of 2e-4, which is adjusted using the \textit{ReduceLROnPlateau} strategy, with a patience of 3, a reduction factor of 0.8, and a minimum learning rate of 1e-8. For the loss functions, Focal loss is applied for confidence estimation, and SmoothL1 loss is used for regression tasks, as in PointPillar \cite{Lang2019}.



\subsection{Results and Discussion}

\subsubsection*{Camera-based Detection}
In this section, we evaluate our proposed feature extractor with different backbones. Similar to our model, the backbones received a single image as input and predicted the 3D characteristics of vehicles with the same HEAD (Section \ref{img2bev}). Furthermore, our evaluation extends to comparing the effectiveness of our model against the car detection capabilities of Monocular 3D Object detection models found in the literature. 

\begin{table}[h]
\caption{Characteristics of different 3D Monocular car detection on the KITTI validation set.}
\centering
\vspace{0.1cm}
\label{table:litmono}
\scalebox{0.8}
{
\begin{tabular}{l|rrccrccc}
\multicolumn{1}{c|}{Models} & \multicolumn{1}{c|}{File size} & \multicolumn{1}{c|}{Params} & \multicolumn{1}{c|}{GPU allocation} & \multicolumn{1}{c|}{Inference time} & \multicolumn{1}{c|}{GFlops} & \multicolumn{1}{c|}{mIoU} & \multicolumn{1}{c|}{mAP(\%)} & Recognition(\%) \\ \hline
MobileNetV3 Small \cite{Howard2019} & 49.101 MB & 12.5 M & 1.1 GB & 43 ms & 134.07 & 0.1687 & 05.20 & 37.41 \\ 
MobileNetV3 Large \cite{Howard2019} & 71.002 MB & 18.0 M & 1.3 GB & 49 ms & 141.32 & 0.2496 & 08.84 & \textbf{69.53} \\ 
ResNet50 \cite{He2016} & 190.605 MB & 26.0 M & 1.6 GB & 53 ms & 161.85 & 0.2116 & 09.73 & 41.32 \\
MobileNetV2 \cite{Sandler2018} & 79.618 MB & 20.2 M & \textbf{0.9 GB} & \textbf{40 ms} & 147.36 & 0.2333 & 11.02 & 52.87 \\ \hline
\textbf{NextBEV (Ours)} & \textbf{5.139 MB} & \textbf{1.2 M} & 1.1 GB & 41 ms & \textbf{71.81} & \textbf{0.2519} & \textbf{13.41} & 54.55
\end{tabular}
}
\end{table}

In Table \ref{table:litmono}, we present a comparison between popular feature extractors and NextBEV. The NextBEV stands out by being a lightweight model with reduced complexity, requiring only 1.2 Million parameters (less than 10\% of MobileNetV3 architecture). This reduction in resource demand, coupled with its lower GPU memory allocation of 1.1 GB, positions NextBEV as a highly efficient model, suggesting potential for deployment in onboard environments without sacrificing computational capabilities. How NextBEV was inspired by MobileNet models, we can observe a similarity in inference time and GPU allocation.

Despite NextBEV showing a significant reduction in parameters and file size, the inference time did not decrease proportionally. We can correlate the number of parameters with the model's file size. However, besides contributing, these parameters are not directly related to the inference time. Some layers, such as Depth-wise Convolution, present more complex processes, which affect the number of parameters more significantly than the inference time. In spite of this, NextBEV achieves the highest scores in key evaluation metrics, leading with a mean IoU of 0.2519 and a mean Average Precision (mAP) of 13.41\%, thus demonstrating its efficacy in recognition tasks with a 54.55\% success rate. As a result, NextBEV indicates a balance between efficiency and accuracy, positioning itself as a potential tool for sensor fusion models that aim real-time applications.

\begin{table}[h]
\caption{Accuracy of 3D Monocular car detection on the KITTI validation set.}
\centering
\resizebox{0.4\linewidth}{!}{
\begin{tabular}{l|cccc}
              & \multicolumn{4}{c}{$AP_{3D}$(IoU\textgreater{}0.7)(\%)}                                       \\ \hline
\multicolumn{1}{c|}{Models}        & \multicolumn{1}{c|}{Easy} & \multicolumn{1}{c|}{Moderate} & \multicolumn{1}{c|}{Hard} & Average \\ \hline
RTM3D  \cite{Li2020}        & 14.41                     & 10.34                         & 08.77                     & 11.17   \\
M3D-RPN \cite{Brazil2019}      & 14.53                     & 11.07                         & 08.65                     & 11.42   \\
MoVi3D \cite{Simonelli2020}        & 14.28                     & 11.13                         & 09.68                     & 11.70   \\
MonoPair \cite{Chen2020monopair}     & 16.28                     & \textbf{12.30}                         & 10.42                     & 13.00   \\
MonoNext \cite{Pederiva2023}     & \textbf{20.76}                     & 09.79                         & 08.70                     & 13.08   \\ \hline
\textbf{NextBEV(ours)} & 19.58                     & 09.17                          & \textbf{11.50}                      & \textbf{13.41}  
\end{tabular}
\label{tab:mono_compair}
}
\end{table}

Besides standing out among the backbones in the literature, we compare NextBEV with 3D Monocular Object Detectors. As showed in Table \ref{tab:mono_compair}, our model demonstrates competitive performance, scoring 19.58\% (Easy), 9.17\% (Moderate), and 11.50\% (Hard), with an average of 13.41\%. The performance is especially noteworthy compared to MonoNext, as our model achieves similar accuracy with fewer parameters and faster inference time.

\subsubsection*{LIDAR-based detection}
To evaluate the impact of our modifications on the performance of LIDAR-based detection models, we conducted a comparative analysis with the default versions of the PointPillar and VoxelNet architectures. In this evaluation, we considered their number of parameters, computational demand, speed, and accuracy. 

\begin{table}[h]
\caption{Comparative analysis of PointPillar and VoxelNet models with our respective proposed versions.}
\centering
\resizebox{0.5\linewidth}{!}{
\begin{tabular}{l|cccc}
                  & Params & GFlops & mAP   & Inference time \\ \hline
PointPillar \cite{Charles2017}& 4.7 M  & 138.09 & \textbf{71.31} & 80 ms          \\
PointPillar Light & \textbf{1.0 M}  & \textbf{75.93}  & 67.27     & \textbf{70 ms}          \\ \hline 
VoxelNet \cite{Zhou2018} & 6.4 M  & 157.64 & \textbf{40.51}     & 105 ms         \\
VoxelNet Light    & \textbf{2.0 M}      & \textbf{94.48}      & 38.88     & \textbf{99 ms}             
\end{tabular}
}
\label{table:lidar}
\end{table}

In Table \ref{table:lidar}, the updated PointPillar model, named PointPillar Light, achieves significant reductions in complexity. This lighter version requires only 1.0 million parameters and 75.93 GFlops, lowering computational demands while maintaining an approximated accuracy. Furthermore, the lighter version reduces the inference time by 10 ms, an essential factor for seeking real-time processing in autonomous vehicles. Similarly, the VoxelNet update demonstrates substantial efficiency gains. The parameter count is less than half of the original, decreasing to just 2.0 million, and GFlops are significantly reduced to 94.48. Like PointPillar, the modifications in VoxelNet reduce its inference time without substantially compromising the model's original accuracy.

\subsubsection*{Sensor-Fusion Detection}


In Table \ref{table:fusion}, we present the comparison of the original LIDAR-only models, PointPillar and VoxelNet, alongside their respective lightweight versions integrated with NextBEV. Noticeably, the lightweight LIDAR versions fused with NextBEV still demonstrate a reduced parameter count compared to the original LIDAR backbone models. The PointPillar Light + NextBEV model, with a reduced parameter set of 3.5 M, exhibits an increase in F1-Score, escalating from 0.4950 to 0.6941. On the other hand, VoxelNet + NextBEV led to more significant advances. Although the number of parameters decreased by 2M, the fusion achieves a mAP of 43.44 \%, and increases the F1-Score in 0.1. Such improvements show that fusion approaches lead to an increase in robust precision-recall balance without compromising the size of the models.

\begin{table}[h]

\centering
\caption{Comparative analysis of default LIDAR-only models and our sensor fusion proposal.}
\resizebox{0.7\linewidth}{!}
{

\begin{tabular}{l|c|cccc}
                            & Sensors-based & Params & mAP   & Infer time & F1-Score \\ \hline
PointPillar \cite{Charles2017}& LIDAR & 4.7 M  & \textbf{71.31} & \textbf{80 ms}          & 0.4950   \\
PointPillar Light + NextBEV & LIDAR+Camera & \textbf{3.5 M}  & 70.28 & 97 ms          & \textbf{0.6941}   \\ \hline
VoxelNet \cite{Zhou2018} & LIDAR & 6.4 M  & 40.51 & \textbf{105 ms}         & 0.3107   \\
VoxelNet Light + Next BEV & LIDAR+Camera   & \textbf{4.4 M}      & \textbf{43.44}     & 125 ms              &   \textbf{0.4152}      
\end{tabular}
}
\label{table:fusion}
\end{table}

\section{Conclusion}

In this research, we present a comprehensive approach to refining the sensor fusion architecture for 3D Object Detection, focusing on enhancing both the image and the LIDAR feature extraction pathways. Our novel architecture, NextBEV, simplifies the transformation of camera images into Bird’s Eye View (BEV) features, achieving high accuracy while significantly reducing the computational overhead. Additionally, we have updated the design of well-known LIDAR backbones to produce models that are both lighter and faster while holding precise detections. By integrating these improvements into a unified sensor fusion architecture, we demonstrate superior reliability and accuracy compared to traditional single-sensor methods. These enhancements mark significant progress toward achieving efficient and high-speed detections necessary for autonomous systems.

\acknowledgments
This study was financed in part by the Coordenação de Aperfeiçoamento de Pessoal de Nivel Superior – Brasil (CAPES) – Finance Code 001, and supported by the Fundação de Apoio ao Ensino à Pesquisa (FAEPEX), State University of Campinas.
\bibliography{report} 
\bibliographystyle{spiebib} 



\end{document}